# Predicting Real-time Crash Risks during Hurricane Evacuation Using Connected Vehicle Data

**Zaheen E Muktadi Syed**
Department of Civil, Environmental and Construction Engineering
University of Central Florida
Email: zaheensyed@knights.ucf.edu

**Samiul Hasan, PhD**
(Corresponding author)
Associate Professor
Department of Civil, Environmental and Construction Engineering
University of Central Florida
Email: samiul.hasan@ucf.edu

## ABSTRACT

Hurricane evacuation, ordered to save lives of people of coastal regions, generates high traffic demand with increased crash risk. To mitigate such risk, transportation agencies need to anticipate highway locations with high crash risks to deploy appropriate countermeasures. With ubiquitous sensors and communication technologies, it is now possible to retrieve micro-level vehicular data containing individual vehicle trajectory and speed information. Such high-resolution vehicle data, potentially available in real time, can be used to assess prevailing traffic safety conditions. Using vehicle speed and acceleration profiles, potential crash risks can be predicted in real time. Previous studies on real-time crash risk prediction mainly used data from infrastructure-based sensors which may not cover many road segments. In this paper, we present methods to determine potential crash risks during hurricane evacuation from an emerging alternative data source known as connected vehicle data. Such data contain vehicle location, speed, and acceleration information collected at a very high frequency (less than 30 seconds). To predict potential crash risks, we utilized a dataset collected during the evacuation period of Hurricane Ida on Interstate-10 (I-10) in the state of Louisiana. Multiple machine learning models were trained considering weather features and different traffic characteristics extracted from the connected vehicle data in 5-minute intervals. The results indicate that the Gaussian Process Boosting (GPBoost) and Extreme Gradient Boosting (XGBoost) models perform better (recall = 0.91) than other models. The real-time connected vehicle data for crash risks assessment will allow traffic managers to efficiently utilize resources to proactively take safety measures.

**INTRODUCTION**

Hurricane evacuation has become a major concern for coastal regions in the United States. With increasing hurricanes, coastal communities need to frequently prepare for mass evacuations—facing an enormous challenge to safely evacuate millions of residents. Potential risk of heavy congestion and increased number of crashes during evacuations prevent local officials to issue evacuation orders. Evacuation creates a surge in traffic volume and the resulting traffic stream follows an oscillatory traffic speed. This may result in crashes especially rear end crashes; previous studies observed a higher number of crashes during evacuation periods (*1*). To ensure safer roads and efficient traffic management of mass evacuations, more proactive and dynamic safety measures should be deployed. To this end, real-time crash risk prediction can enable safer and more efficient evacuations of a high number of vehicles.

With limited resources, taking proactive traffic measures at the right time has become increasingly challenging. It is not enough to mitigate the number of crashes by just identifying high risk zones from historical crash data. Using evacuation data collected during Hurricane Irma, a recent study (*2*) has found that there is a higher crash risk during hurricane evacuations than regular periods. If crash risks can be predicted in real time, traffic management agencies will be able to assess the safety conditions of roadways more proactively and place traffic safety measures more strategically. Consequently, crash risk prediction using real-time traffic data has been a widely investigated topic among traffic safety researchers (*3*). Similarly, researchers (*4*) (*5*) (*6*) have also established strong relationship between vehicular data like speed and acceleration and crash risk potential in different contexts from various data sources.

Connected vehicles, capable of real-time communication using vehicle-to-everything (V2X) technologies such as network-enabled on-board units (OBUs), have enabled us to create data pipelines and facilities to have real-time vehicular data (*7*). This type of micro-level vehicle data consisting of vehicle's positional and speed information is called floating car data points (FCD) (*8*) and the vehicles are called as connected vehicles. Previous research programs such as Strategic Highway Program (SHRP2) collected similar microscopic data on selected vehicles over a specific period. However, with the advent of connected autonomous vehicles (CAVs) in the market and integration of smart technology like OBU units, it is anticipated that these datasets will be more readily available. These connected vehicle data can be used for traffic estimation and prediction purposes. A recent study (*9*) has shown that even with a penetration rate as low as 1.5%, connected vehicle data can be used to predict traffic volume and speed up to 60 minutes ahead with reasonable accuracy levels. Studies have shown that vehicular speed characteristics and acceleration profiles can be used to calculate surrogate safety measures (*10*) which in turn can be used to identify potential crash risks (*11*). In this paper, we have developed a framework utilizing only connected vehicle data to train a machine learning model for predicting potential crash risk in real time during hurricane evacuation.

Many previous studies (*12*) (*13*) have used infrastructure-based data collection methods (e.g., loop detectors and Microwave Vehicle Detection System [MVDS]) to predict real-time crash risks. These predictions rely on sensors leading to issues such as the cost of installing the sensors, lack of coverage, potential data loss due to hardware failure, high maintenance cost, and fixed locations of the sensors. The decentralized nature of connected vehicle data creates an opportunity to assess traffic flow patterns and safety conditions of any road segment without deploying infrastructure sensors. Thus, real-time crash risk prediction relying on connected vehicle data will allow a wider coverage of roads.

The connected vehicle data used in this paper is accumulated from existing onboard devices deployed in connected vehicles by a third-party automobile data service platform (*14*). The data collected is further processed to determine the potential crash risk of interstate road segments. However, care is taken such that post processing does not introduce any latency in the system. Our work is independent of roadside detectors or any infrastructure dependent data collection method and therefore can be implemented in any highway. This paper contributes in three ways: first, we develop a framework to utilize connected vehicle data to identify potential crash risk; second, we train powerful machine learning models to predict potential crash risks and identify important features; and third, our models are suited to predict crash risk during hurricane evacuation. To the best of our knowledge, this is also one of the first studies that has used the Gaussian Process Boosting model, a powerful model incorporating boosting with Gaussian process and mixed effects models, for real-time crash risk prediction.

**LITERATURE REVIEW**

To understand the influence of driving behavior on crashes, researchers used naturalistic driving behavior data (*4*) collected from GPS sensors, onboard vehicle sensors and video data. Singh et al. (*15*) reviewed naturalistic driving studies and concluded that it is possible to improve driving behavior through providing feedback to the drivers about their driving and surrounding. However, the authors found that extensive data processing, privacy concerns, and low participation of drivers in data collection efforts are major limitations of these studies. On the other hand, previous studies developed models to predict real-time crash risks using a wide variety of data mostly available from infrastructure-based sensors. Hossain et. al. (*3*) reviewed real-time crash risk prediction models and concluded that such models have the potential to identify and intervene in safety hazards in advance. But some challenges remain including the lack of reliable real-time data and appropriate modelling methods. Therefore, exploiting a reliable data source observing naturalistic driving behavior could be gamechanger towards the development and application of real-time crash risk prediction models.

Studies based on naturalistic driving data have extensively investigated the connection between crash risk and driving behavior. Wahlberg (*16*) investigated the relationship between acceleration behavior and crash frequency using local bus data and suggested that the absolute mean of speed changes has a better predictive power than speed. Jun et. el. (*17*) utilized GPS data of 167 drivers over 14 months and observed that speed and acceleration patterns of drivers who were involved in crashes were significantly different to other drivers who were not involved in crashes. The study suggested that vehicle monitoring technologies have a great potential to identify risky driving behaviors. Xie et al. (*18*) analyzed GPS data from taxis and concluded that higher speed was correlated with crash occurrence. Stipancic et al. (*11*) discovered that hard braking events and hard acceleration are positively correlated with crash frequency and severity. Also, Stipancic et. al. (*19*) used hard braking, congestion level, and speed variation to develop a network screening application and found that these features are positively correlated with crash frequency at intersections. Using naturalistic driving data, Kamrani et. el. (2019) (*20*) found that speed will increase the chance of near crash and crash events. Desai et al. (*21*) and Hunter et al. (*22*) also confirmed the positive correlation of hard braking with crash frequency. These studies investigating the relationship between individual driving behavior and crash risk also identified important traffic features that can be used to predict crash risk. As the availability of vehicular data has increased, researchers have started to investigate if real-time crash risks can be predicted models from such data.

Previous studies on real-time crash risk predictions have predominantly used infrastructure-based sensors as a reliable data source. For instance, to predict real-time crash risks, Yu and Abdel-Aty (*23*) used 5-minute aggregate data from Microwave Vehicle Detection System (MVDS) detectors recording speed, volume and occupancy. To predict crash risk, they trained a Support Vector Machine (SVM), a lightweight powerful machine learning model and discovered that traffic features such as downstream average speed, average speed, and standard deviations of occupancy and volume at crash locations are the most important features for predicting crash risk. Recently, a matched case control-based approach has been used to identify the factors contributing to the increased number of crashes during evacuation (*12*). The study used MVDS detector data (including traffic speed, volume, and occupancy) to estimate traffic attributes upstream and downstream of a crash location. Their model results show that the likelihood of crash occurrence increases with a high volume of traffic at an upstream segment and a high variation of speed at a downstream segment. Wang et al. (*24*) used a random forest algorithm to find the important features influencing crash prediction and showed that traffic variables have a higher ranking in terms of importance followed by weather data and then spatial features like ramp geometry. Thus, most of this research have proved that aggregated traffic data have the potential to predict crash risk accurately and reliably in real time, but all these works rely on infrastructure-based sensors which have limited coverage in transportation systems.

Another major aspect of real-time crash risk prediction is the selection of an efficient method that can be deployed in real time. In recent years, machine learning models have been widely applied due to their predictive power and ease of application. Popular machine learning methods include support vector machine (*23*), random forest (*25*), eXtreme gradient boosting (*26*), and various neural networks etc. Wang et al. (*27*) used Adaptive Boosting and Support Vector Machine to identify incidents from large traffic data collected from signal video data. Huang et al. (*28*) used traffic detector data to develop a convolution neural network (CNN) to detect crash occurrence and estimate crash-prone traffic conditions. Li et al. (*29*) developed a deep learning architecture called LSTM-CNN model to predict crash risk in urban arterials. Overall, machine learning models have proved to be a reasonable choice for predicting real-time crash risks. As such, recent research has started using alternative data sources and machine learning methods to develop real time crash risk prediction models. For example, Shile et al. (*30*) developed a bidirectional LSTM model to predict crash risk on freeways using connected vehicle data and detector data together. However, some machine learning methods such as neural network based models would require large-scale data to train them.

Predicting real-time crash risks during a hurricane evacuation can be challenging as evacuations typically last two to five days leaving limited data to train complex machine learning models. In addition, evacuations may generate unpredictable traffic conditions that may contribute to crashes. To confirm this, Rahman et al. (*2*) used a matched case control approach to analyze MVDS data collected during Hurricane Irma and discovered increased crash risks during evacuation compared to regular periods. Xu et al. (*32*) investigated the relationship between traffic flow and crash risk under different traffic conditions and suggested that the factors contributing to crash risks are quite different across various traffic states. As such, traffic conditions generated in a congested highway during a hurricane evacuation are likely to be different from normal times (*1*)—requiring crash prediction models to be developed based on real-world evacuation traffic data (*31*).

In summary, there is strong evidence that crash risks can be predicted in real time using traffic features. Previous studies have mostly used infrastructure-based data sources with limited traffic features (such as average speed, volume, and occupancy) extracted from those data. However, these data sources mostly rely on infrastructure sensors with limited coverage and aggregate features. Utilizing connected vehicle data, we can determine various useful features such as maximum speed, acceleration rate, deceleration rate and standard deviation of speed which reflect individual driving behaviors. Similar to the recent studies on real-time crash risk prediction, we adopted machine learning based models. Since we were interested in predicting real-time crash risks during evacuation, we focused on shallow machine learning models that would require small samples of the data. Despite the data size, our analysis results show that we can predict real time crash risk with very high accuracy.

**DATA DESCRIPTION**

Hurricane Ida, a category 4 hurricane, made its landfall on 29$^{th}$ August 2021 in the coastal regions of Louisiana causing more than 75 billion dollars in damages (*33*). Due to the rapid intensification of the hurricane, authorities were able to predict its potential impact only two days before the landfall. The state issued a state of emergency on 27$^{th}$ August 2021 and ordered mandatory evacuation of coastal counties near

New Orleans. Figure 1 is a parish map of the Louisiana state, and all the blue colored parishes are coastal regions. Evacuation orders were placed in the dark blue regions (parishes) which were anticipated to be near the path of the Hurricane Ida, denoted by the yellow line. The I-10 interstate (denoted by the red line in Figure 1) is the major highway going across the state of Louisiana. It is expected that during the evacuation period, I-10 would face high traffic volume. We selected the 291 miles of I-10 road segment as our study area, shown in Figure 1. We divided our study area into 124 segments in both directions. Each segment is about 2.25 miles long, the segmentation is done based on the latitude and longitude value of the interstate highway, therefore the length of the segment varies with a standard deviation of 0.91 miles.

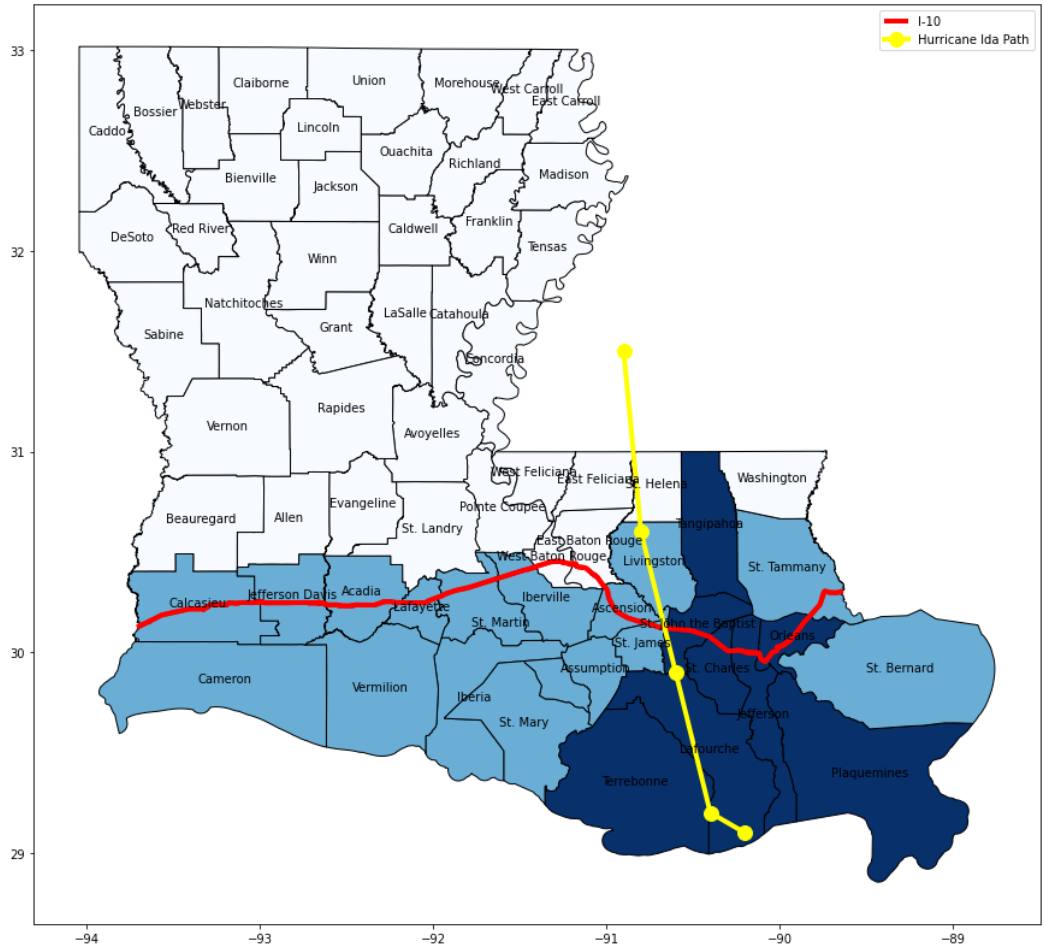

*Figure 1: Louisiana Parish map showing the study area and Hurricane Ida path.*

In this study, we utilized three types of datasets: connected vehicle data, crash data, and weather data. Many companies have ventured in compiling connected vehicle data in real-time and gather traffic insights. We procured data from a vehicle data provider, Otonomo (*34*), which specializes on connected vehicle data. We opted to use 3 days data for the selected segments.

The raw data consisted of speed, acceleration, and location of anonymous vehicles collected at a frequency less than 30s. The initial dataset had more than 5 million datapoints. We extracted the information of 31,186 unique vehicles that were driven on I-10 in the 3 days of evacuation period of Hurricane Ida. The data were segregated for each road segment into 5-minute time intervals. For each day, there are 288-time intervals, but we removed the time intervals for which no observations were available. Then, for each segment and time interval, we calculated a range of features as listed in Table 1 based on the vehicular data.

In this study, we opted to use only real-time features to predict potential crash risk. Previous studies (*35, 36*) proved that traffic features such as speed variation and acceleration profile measurements are strongly associated with crash risks and therefore can also be adopted as surrogates for safety measures. Apart from different speed related features, we also extracted a few features related to acceleration as previous studies (*11*) found that vehicle maneuvers such as hard braking and acceleration events are strongly correlated with crash occurrence. Thus, based on previous literature, we considered most of the real-time traffic features that can be measured and calculated from connected vehicle data. Table 1 lists all these features gathered from reviewing previous literature along with mean, standard deviation, minimum and maximum values of each feature in the final processed data. Some features required threshold values. We used previous literature and traffic knowledge to intuitively set those threshold values. However, in this table, we only present the core features but in our final dataset also included features to account for traffic conditions in upstream segment, downstream segment and previous three-time intervals. The detailed process of feature extraction and calculation is described in the next section.

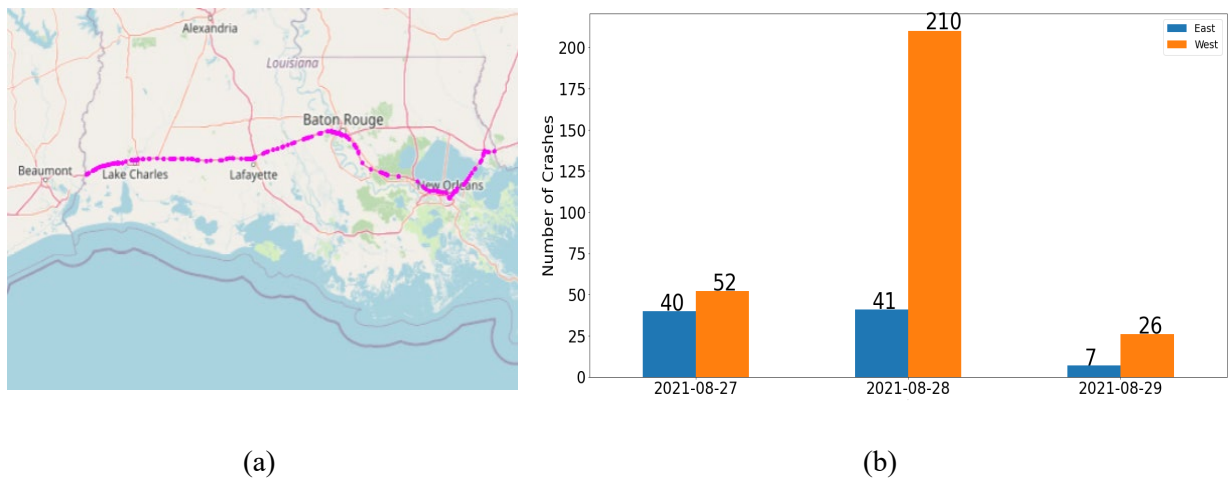

(a)　　　　　　　　　　　　　　　　　　(b)

*Figure 2: Crash Distribution in I-10 (a) spatial distribution (b) numerical distribution*

We also used a separate weather data source to record weather features. We retrieved the nearest airport weather station to collect the hourly weather information and merged the weather data with our analysis. Finally, we merged the crash data information from RITIS (*37*) by aggregating all the traffic crashes on I-10 during the evacuation period. We found 376 crashes, 288 in the westbound direction and 88 in the eastbound direction, during the 3 days of evacuation on our selected study area. The crash points are plotted in a map and shown in Figure 2(a) and the number of crash in each direction for each day is plotted in bar graph shown in Figure 2(b). If a crash occurs at any designated segment, we will flag the time interval of that segment with the numeric digit 1.

Table 1: Feature names and description

| Data Source | Feature Name | Feature Description | mean | std | min | max |
|---|---|---|---|---|---|---|
| Crash data | crash_check (target variable) | 1- if a crash occurred for a given segment in each time interval; 0- otherwise (i.e., no crash occurred) | 0.002 | 0.045 | 0.0 | 1.0 |
| Connected vehicle data | acc_cal | Calculated acceleration ($m/s^2$) | 0.302 | 0.424 | 0 | 8.899 |
| | acc_cnt_thres | Number of vehicles accelerating above threshold (3.4 $m/s^2$) | 0.001 | 0.04 | 0 | 5 |
| | dcc_cal | Calculated deceleration ($m/s^2$) | 0.439 | 0.578 | 9.836 | 0 |
| | dcc_cnt_thres | Number of vehicles decelerating above a threshold (3.4 $m/s^2$) | 10.058 | 20.675 | 0 | 352 |
| | max_acc | Maximum value of the acceleration recorded ($m/s^2$) | 0.242 | 0.490 | 0 | 7.042 |
| | max_dcc | Maximum of the absolute value of deceleration recorded ($m/s^2$) | 0.086 | 0.227 | 0 | 4.081 |
| | max_speed | Maximum speed ($km/hr$) | 112.278 | 28.955 | 0 | 260.713 |
| | mean_speed | Mean speed ($km/hr$) | 96.774 | 35.489 | 0 | 222.089 |
| | speed_cnt_thres | Number of records above speed threshold of 160.9 km/hr (100 mph) | 18.899 | 22.415 | 0 | 312 |
| | ss | Standard deviation of speed | 9.459 | 9.578 | 0 | 79.821 |
| | timeofday | Time of the day in hours | 12.084 | 7.018 | 0 | 23 |
| | up_time_diff | Time difference between upstream and current segment interval beginning times. | 12.084 | 7.018 | 0 | 23 |
| | veh_cnt | Number of distinct vehicles observed in the data | 2.091 | 1.579 | 1 | 16 |
| Map | dist_seg | Length of the segment (miles) | 2.259 | 0.919 | 0.00497 | 2.972 |

| Weather | HourlyDryBulbTemperature | Temperature in Fahrenheit | 78.902 | 4.704 | 71 | 89.33 |
| | HourlyPrecipitation | Precipitation measure (mm) | 0.901 | 3.305 | 0 | 68 |

## METHODOLOGY

### Data Processing

The raw connected vehicle data consists of individual microscopic vehicle data i.e., time of record, vehicle location, vehicle speed and acceleration. As we were interested to understand the crash risk for each road segment, we processed this data so that it reflects the traffic features at the level of a road segment. First, we checked the raw data for any unreasonable values. For example, if any speed is negative and to detect any outlier in the speed using an interquartile range. We calculated the 25% quantiles (Q1) and the 75% quantiles of the feature value (Q3). The interquartile range, Q4, is calculated by subtracting Q1 from Q3, i.e., Q4=Q3-Q1. We use the interquartile range to calculate the upper bound and lower bound. For speed, we only calculated the upper bound which is 245.4 km/h (upper bound: $Q3 + 1.5 * Q4$ ) and kept the lower bound as zero. For acceleration, we used a threshold to detect outliers. The threshold in this case was set at $\pm 13\ m/s^2$ for acceleration and deceleration respectively as this is the maximum acceleration achieved by Tesla Model S plaid, the fastest production car by acceleration (*38*, *39*). The distributions of the speed ($km/h$), acceleration ($m/s^2$) and deceleration ($m/s^2$) are plotted in Figure 3.

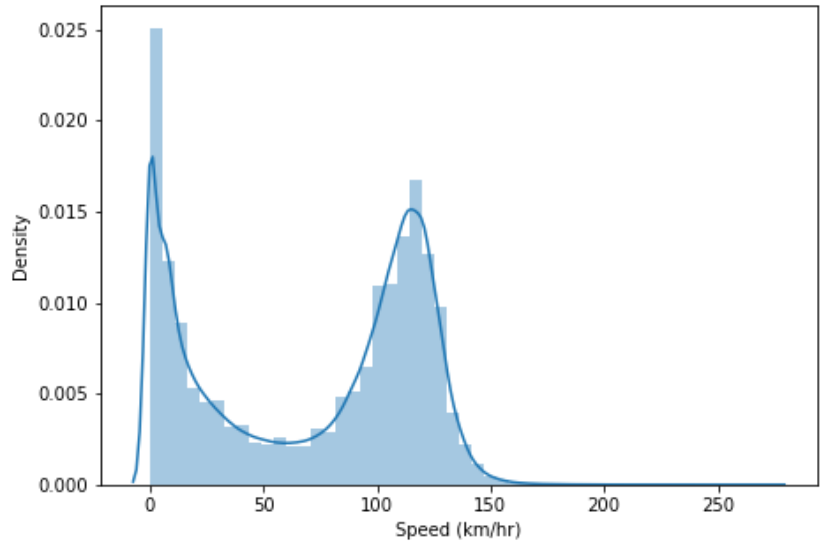

(a) Speed distribution

Syed and Hasan

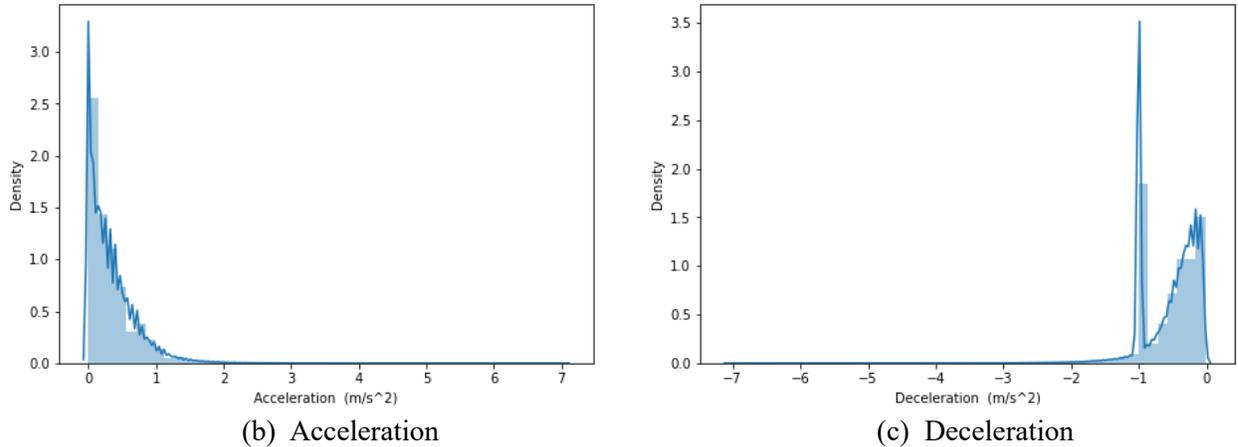

(b) Acceleration  (c) Deceleration
Figure 3: Distribution of (a) speed in km/hr (b) acceleration in $m/s^2$ and (c) deceleration in $m/s^2$

Then the clean data was passed through segmentation, where we separated the data based on their spatial information and the pre-determined segments of the study area. The predetermined segments are flexible, and we can control the segment length. After spatially locating the data points, we aggregated all the data of each segment into five-minute time intervals for each segment. Segregating the data into 2-mile segments and aggregating the data in 5-minute interval is done to make our results comparable to previous literature. We observed that most previous works used detector data, and the average distance between two detectors varied from 0.5 mile to 2 miles (*13*, *40*). So, we decided to take the higher range of 2 miles or more for this analysis. Also, most study used detector data aggregated at a 5-minute interval (*23*, *29*, *41*). Finally based on the aggregated data, we calculated each feature as referred in Table 1.

Most features are straight forward, and common traffic terms. Co-efficient of variation of speed (CVS) was initially calculated but removed later as CVS was calculated from mean speed divided by standard deviation of speed, both of which are already present as input features. Two features, acc_cal and dcc_cal were calculated based on the speed of individual vehicle in each segment and time interval. From the trajectory of each vehicle, the acceleration and deceleration are calculated using the simple formula as shown in Equation 1. The mean of acceleration and deceleration is then recorded as acc_cal and dcc_cal respectively for each segment and interval of time. This is because the acceleration data provided may have lot of discrepancies which can be introduced from sensor vibration, but speed data is more reliable and accurate.

$$a = (v_1 - u_1)/(t_d) \quad (Equation\ 1)$$

We created three features where the number of vehicles is counted based on threshold values, i.e., speed_cnt_thres, acc_cnt_thres and dcc_cnt_threshold. In our feature calculation, we used 160.9 km/hr (100 mph) as our speed threshold to capture the super speeding vehicles which is almost 1.5 times the speed limit of 112.7 km/hr (70 mph). From our distribution of speed as shown in Figure 3, we observe two peaks one of which is close to 0 km/hr and another one is at around 125 km/hr (77 mph). This means a majority of the vehicles either travelled at very low speeds and others traveled at speed higher than the speed limit. In Florida, where the study is done, driving 30 mph or more above the speed limit is considered super speeding and imposes the highest speeding ticket by law. Thus, to capture the super speeding vehicles, which tend to pose greater risk of crash, we have set the speed threshold 1.5 times the speed limit (30 mph higher the speed limit). We set $\pm 3.4\ m/s^2$ as our threshold for acceleration and deceleration, respectively. AASHTO recommends (*42*) a deceleration rate of 3.4 $m/s^2$ (11.2 $ft/s^2$) when calculating stopping sight distances; such a deceleration rate is within the capability of 90% of the drives as found in empirical studies (*43*), but would indicate an unsafe driving condition (*42*). Furthermore, simulation-based studies (*1*) using

Syed and Hasan

deceleration rate to avoid collision (DRAC) set as $3.4\ m/s^2$ to detect a crash event. (*43*). Thus, the acceleration and deceleration thresholds are set at $\pm 3.4\ m/s^2$.

The features listed in Table 1 show only the features calculated for a given segment and current time interval. Features were also calculated to reflect traffic conditions at upstream ($u_1$), downstream ($d_1$) and previous time intervals ($t_1, t_2, and\ t_3$). Previous studies (*23, 29*) found that these features are indicators of crash risk. For each segment, we input 5-, 10- and 15-minutes of previous data of all traffic features, denoted by $t_1, t_2, and\ t_3$, respectively. However, not all the features will be in the final model, only important and uncorrelated features will be included.

Weather features, as shown in Table 1, are also checked for outliers, segmented based on the closest weather station and recorded for each time interval. The weather data is updated every hour and therefore may not be very accurate. In summary, the data preparation process is shown in Figure 4. Data from connected vehicle data and weather data are first cleaned and then segmented, aggregated, and used for feature calculation to produce the final prepared data.

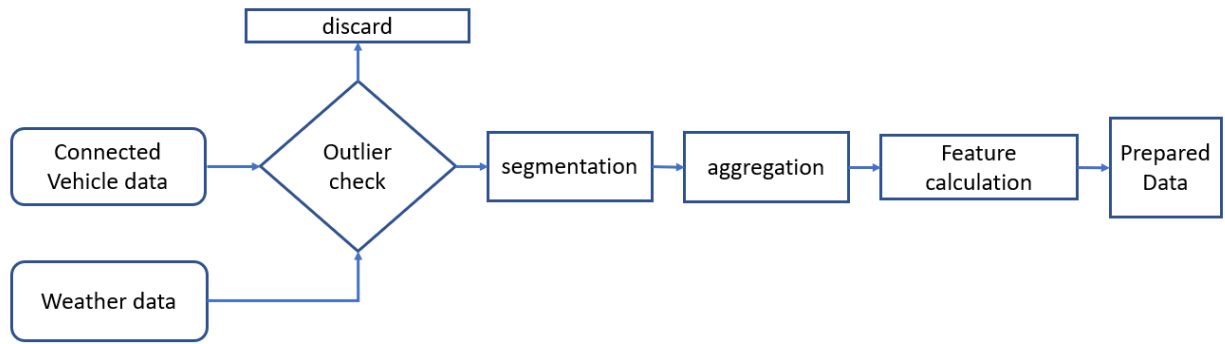

Figure 4: Data preparation framework

**Data Analysis**

The connected vehicle raw data can give us insights on the level of speed variation for any road segment. Unlike detector data, the processed features account for any irregular driving behaviors that can lead to higher crash risks. To demonstrate this, we plot the mean speed vs. time graph for one road segment over all time intervals, as shown in Figure 5. The graph shows the speed characteristics of one random road segment before and after a crash event (marked by the purple vertical line). The shape of the graph clearly shows a high variation of speed just before the crash event and then a gradual decrease in the speed after the crash event. (*11*)

Syed and Hasan

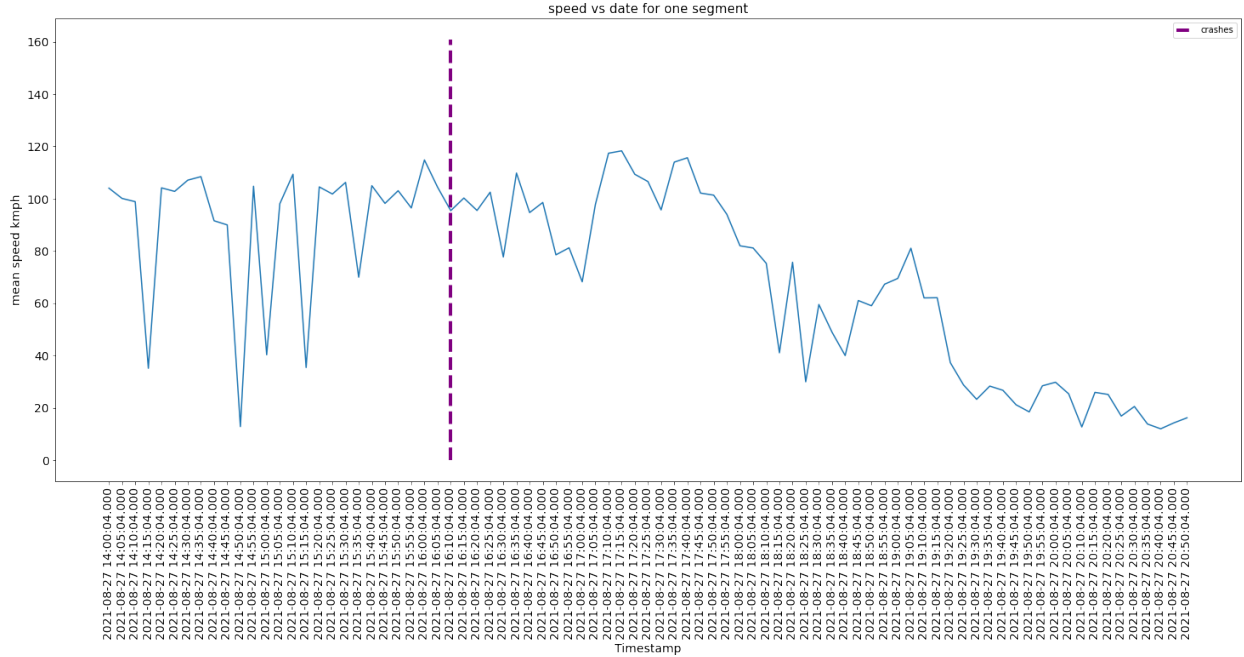

Figure 5: Speed vs. time for a road segment before and after a crash event

To ensure that each feature is independent, we plotted the correlation matrix. A portion of the matrix with all core features are shown in Figure 4. Most features have very low correlation (less than 0.5) between each other except for the max speed and mean speed. The importance of each feature is determined before deciding the final features of the selected model.

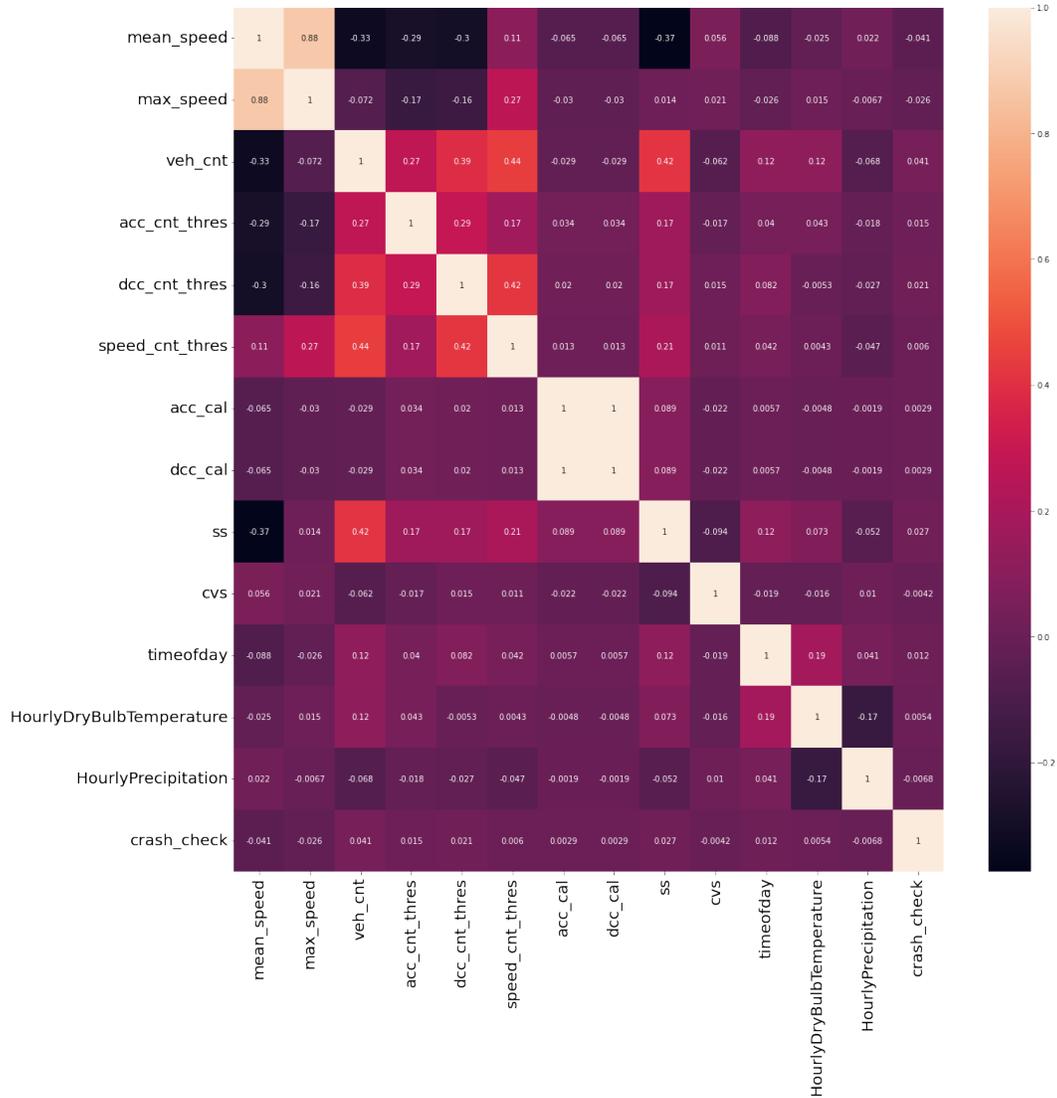

Figure 6: Correlation matrix of the traffic features

**Data Balancing Method**

A common problem for crash prediction is that the data is highly imbalanced. For instance, after the final dataset is processed, we are left with only 180 crashes and 88,060 non-crash instances. This high difference between the numbers of crashes and non-crash events leads to bias and deteriorates the performance of the selected machine learning models. To overcome this issue, previous studies have used case control matching (*2*, *44*, *45*) and sampling techniques (*46*). Popular data balancing methods used for crash risk prediction include synthetically producing data using over sampling methods such as SMOTE (Synthetic Minority Oversampling Technique) (*47*) and Deep Convolutional Generative Adversarial Network (DC-GAN) (*13*). Another important reason for checking the correlation between data features is that while synthetically producing crash data the correlation of the features should be considered.

In this study, we used the SMOTE technique (*48*) to make our results comparable to previous literature. Each instance of the minority samples (in our case the crash events) has a vector created by SMOTE between it and its *k*-nearest neighbors (in the same dataset). Then, each vector is multiplied by a chance constant between 0 and 1 to create a new example data vector, which is then added to the dataset. The operation is repeated until the desired sampling rate is reached.

**Crash Prediction Models**

Since our target variable is binary (i.e., whether or not a crash event has occurred), our problem becomes a typical classification problem. Initially, we applied Logistic Regression, Support Vector Machine (SVM), and Random Forest models that have been commonly used for crash risk prediction purposes. Besides, we have applied a Gaussian Process Boosting that combines boosting with Gaussian process and mixed effects models (*49*). We have also applied state-of-the-art XG boosting on the original dataset.

*Logistic Regression*

The logistic regression model has been commonly used in transportation safety research (*50*). The model serves as a baseline model for comparing the performance of other models. Since it is widely used, the results can validate the proposed use case of the connected vehicle data. To fit the model, we converted our dependent variable, crash risk, into binary indicator of crash occurrence, where the model calculated the probability $p$ for crash case ($y = 1$). We have implemented logistic regression using the regularized logistic regression classifier module of the Sklearn library (*51*). We used the default parameters and varied the maximum iteration parameter which controls the maximum number of iterations before which the solver converges.

*Support Vector Machine*

The Support Vector Machine (SVM) is another powerful classifier used to solve binary classification problems. SVM has also been adopted to predict crashes (*52*), including real-time applications (*23*). From earlier implementation of SVM in real-time crash prediction, it has been proven to be effective in detecting crash risk with low misclassification rate, low false alarm rates and higher accuracy. We implemented SVM as a binary classifier using the linear SVC class of the Sklearn library. The linear SVC is a faster implementation of support vector machine used for classification. For SVM, we also used the default parameters, but only varied the maximum number of iterations.

*Random Forest*

The Random Forest (RF) is an ensemble learning method that fits a number of decision tree classifiers on various sub-samples of a dataset and considers the average of their predictions to improve predictive performance. The various subsamples are generated using different validation strategies like bootstrap and bagging. The RF model can be used for both classification and regression problems; it is relatively fast and efficient to implement. RF model is commonly used for feature importance selection and rank the important features for real-time crash prediction (*53*). We used the random classifier module of the Sklearn library, with the default parameters. We calculated the feature importance from the best model results.

*Gaussian Process Boosting (GPBoost)*

Boosting is a common machine learning technique that can achieve high predictive performance for a wide variety of datasets (*54*). Gaussian Process Boosting (GPBoost) integrates boosting with Gaussian process and mixed effects models. Boosting, a powerful machine learning technique with high accuracy, assumes independence among features and may deteriorate probabilistic prediction for data that may have random effect. Latent Gaussian models, encompass Gaussian process and mixed effect models, can be used to model dependence among features and make probabilistic prediction (*49*). Since the road segments are spatially distributed one after another, we can assume that traffic features of one segment can affect the crash risk of the surrounding segments. Thus, there may exist some unobserved heterogeneities across various road segments that may affect crash risk. This is commonly referred as random effects (*55*, *56*). As GP boosting incorporates both tree-boosting and gaussian process, we can also minimize any effect of multicollinearity that may have been introduced by the road segment or from the calculated features. Using GP boosting model, we can also consider the random effects that may exist in our data and predict potential crash risk with higher accuracy than corresponding statistical models.

*Extreme Gradient Boost (XGBoost)*

Over the years, many variations of the boosting techniques are developed, each utilizing an ensemble of relatively weak prediction models like decision tree to obtain a better prediction model. Gradient Boosting, also a variant of boosting models, uses gradient descent method to add weak classifiers and minimize the loss of the model. The loss function depends on the type of the problem. Since our problem is a classification one, we used the logarithmic loss function. Extreme gradient boosting uses the same concept of Gradient boosting except it is more adjusted to minimize the loss function and prevent underfitting or overfitting. XGBoost has the potential to outperform ensemble of neural network (*54*). We used the XGboost library for implementation and we only changed the maximum depth and learning rate of the gradient descent. From a grid search, we found a learning rate of 0.05 and a maximum depth of 11 as the best performers.

**Model Data Preparation**

For modeling, the processed data had to be split into a training dataset (70%) and test dataset (30%). Since our dataset was highly imbalanced, we had to be very careful while splitting the dataset so that no dataset is devoid of any crash cases. To split the dataset, we separated the crash cases and non-crash cases. Then we used Sklearn test train split function to randomly divide the respective dataset into test (30%) and train dataset (70%). We now have four datasets, two from crash and two from non-crash. Next, we concatenated the training crash cases and training non-crash cases datasets into one final training dataset. Similarly, we concatenated the crash testing set and non-crash testing dataset together. However, both the datasets were still imbalanced. Then, we applied the SMOTE technique to balance the crash and non-crash cases of each dataset. This process ensures that both final train and test datasets have crash and non-crash cases and random selection of instances to avoid model overfitting.

**Feature Importance**

To understand the importance of each feature, we ranked the features in two ways: first we calculated the feature importance from Random Forest model using permutation feature importance and second, we used Shapley Additive Explanations (SHAP) feature importance (*57*) calculated from the XGBoost model. Permutation feature importance measures the decrease in a model score by randomly shuffling a single feature. The Shapley feature importance is calculated from the shapely value, widely used in game theory, which is the average of the expected marginal contribution of one feature after all possible combinations have been considered. The main difference between SHAP and permutation is that SHAP assigns each feature an importance value based on the feature's contribution toward its prediction while permutation calculates the importance of a feature based on the decrease in model performance. Thus, using permutation, we can identify features that are negatively impacting our prediction.

From the Random Forest model, we ranked the most important features and identified the features that had negative feature importance. A negative feature importance means that the variable does not have a role in prediction and randomly affect the predicting variable. Initially, we calculated the feature importance using all the features. We found that acc_cnt_threshold, dcc_cnt_threshold, temperature, and coefficient of variation of speed had negative importance values indicating that these features were negatively impacting model performance. Thus, we removed these features from all the model input along with their upstream, downstream, and other time interval counter parts. Finally, we had only 54 features, all of which are plotted in accordance with their importance as shown in *Figure 7* . It is important to note that the figure presented here only shows the final features.

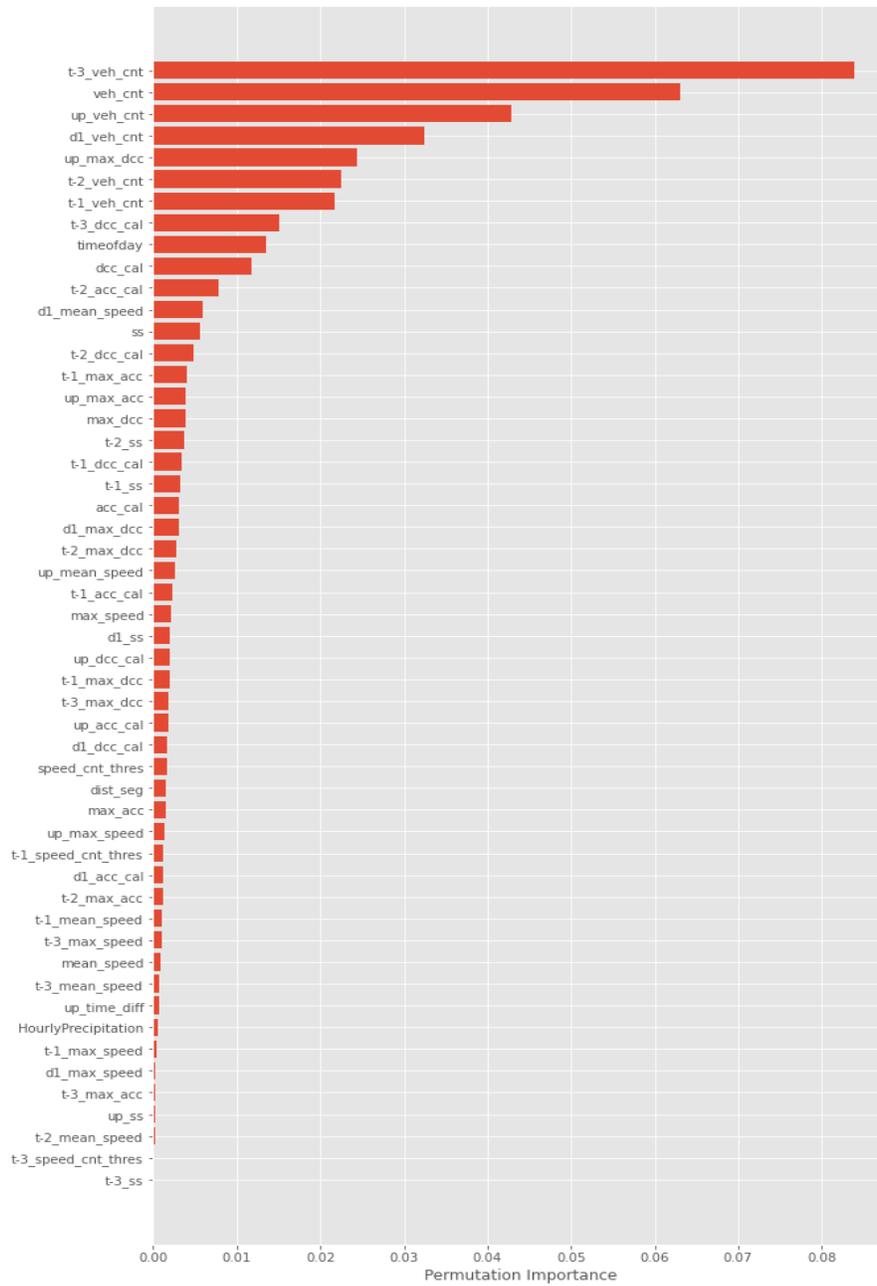

Figure 7: Feature Importance from RF model

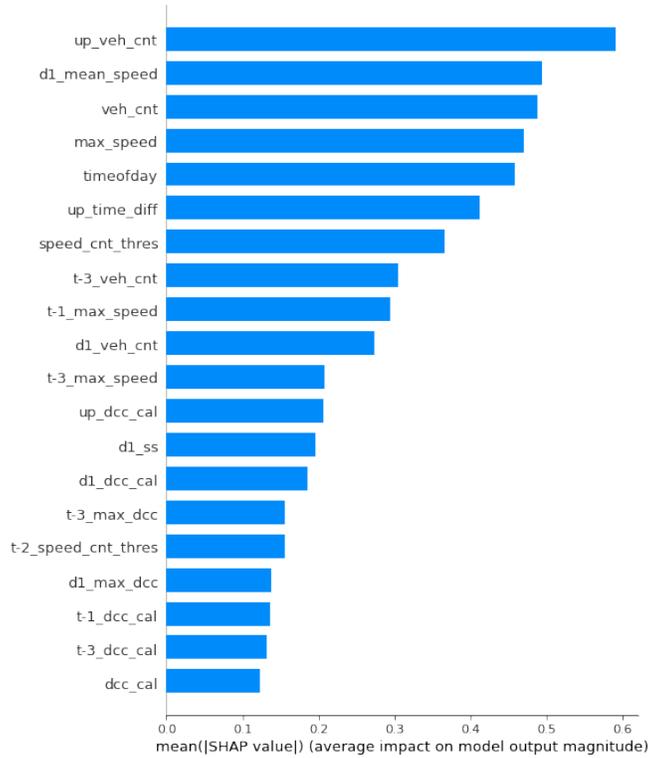

Figure 8: SHAP value-based feature importance from XG Boost model.

After selecting the final features, we ran the XG boost model to calculate the SHAP values. The important features identified from SHAP values are plotted in *Figure 8*. Upstream vehicle count is identified as the most important feature. The vehicle count is representative of the volume of the traffic, and as per both models, vehicle count has one of the highest effects in predicting crashes. Vehicle count can be considered as a proxy of traffic volumes. Besides vehicle count, downstream mean speed, max speed, speed_cnt_threshold, upstream deceleration, and downstream standard deviation of speed are also identified as important features. We also observe that time of the day and the upstream time difference, which are non-traffic features but available in real time, also play an important role in predicting crashes.

**RESULTS**

We have calculated the precision, accuracy, sensitivity, and F1-score to evaluate the results of each classification model. The definition of each evaluation metric is given in Table 2.

Table 2: Evaluation Parameter Definition

| Parameter | Definition |
|---|---|
| Accuracy | Measures the percentage of prediction correctly identified. |
| Precision | Measures how many of the predicted positive classes are correct. |

Syed and Hasan

| | |
|---|---|
| Recall (Sensitivity) | Measures how many of the actual positive classes the model can correctly predict. |
| F1 Score | It is the weighted average of precision and recall. F1 score has a range between 0 (worst case) and 1 (best case). |
| Specificity | Measures the how many of the actual negative classes the model can correctly predict. |

We used a sampling ratio 1:1 (SMOTE 1), similar to the previous studies that used SMOTE. A sampling ratio of 1:1 means that SMOTE will synthetically generate the minority cases needed to make the number of minority (crash) and majority cases (non-crash) to be equal. We also carried out a sensitivity analysis on the sampling ratio of crash to non-crash event as 1:2 (SMOTE 0.5) and 1:4 (SMOTE 0.25).

Each model was run 10 times and the average performance value for each model for each sampling strategy is presented in Table 3. For a crash prediction model, it is important to know how many actual crashes are correctly identified by the model. Thus, the recall is the most important evaluation measure as it indicates what proportion of crashes the model can correctly predict. The base model, logistic regression, shows a recall value of 0.66 for a sampling ratio of 1:1. The SVM and logistic regression models have similar performance in terms of precision and recall across all sampling strategies. However, we observe a significant improvement in precision values for the Random Forest model in comparison to both logistic regression and SVM. A review of real-time crash prediction models by Hossain et al. (*3*) found that only 15% of the previous studies were able to report crash detection rate higher than 81%. Thus, a recall value of 0.81 is better than most models used in real-time crash prediction. With XGboost model we achieved a higher recall value (0.87, 0.88, 0.91) for all respective sampling ratio (0.25, 0.5, 1). The recall result for both XGBoost and GPBoost models at a sampling ratio 1:1 was observed as the highest (a recall value of 0.91). GPboost can be a powerful model as it has the efficiency and scalability of ML boosting model and the capability to handle random effect due to spatial relation of road segments. This makes the model a perfect fit for predicting crashes. Overall, both the XGBoost and GPBoost models have great potential at predicting real-time crash risks. The highest recall rates achieved at each sampling rate is shown in bold (Table 3).

Table 3: Model results on test dataset with different sampling rates

| Event | Model Measure | Sampling (Sampling ratio) | Logistic regression | SVM | Random Forest | XGBoost | GPBoost |
|---|---|---|---|---|---|---|---|
| Crash | Recall / Sensitivity | SMOTE 0.25 | 0.21 | 0.18 | 0.54 | **0.87** | 0.58 |
| | | SMOTE 0.5 | 0.44 | 0.44 | 0.73 | **0.88** | 0.77 |
| | | SMOTE 1 | 0.66 | 0.66 | 0.86 | **0.91** | **0.91** |
| | F1-score | SMOTE 0.25 | 0.30 | 0.26 | 0.69 | 0.93 | 0.73 |
| | | SMOTE 0.5 | 0.53 | 0.53 | 0.82 | 0.94 | 0.84 |
| | | SMOTE 1 | 0.70 | 0.70 | 0.90 | 0.95 | 0.89 |
| | Precision | SMOTE 0.25 | 0.52 | 0.51 | 0.95 | 0.99 | 0.99 |
| | | SMOTE 0.5 | 0.66 | 0.68 | 0.93 | 0.99 | 0.94 |
| | | SMOTE 1 | 0.75 | 0.75 | 0.94 | 0.99 | 0.87 |
| | Specificity | SMOTE 0.25 | 0.95 | 0.96 | 0.99 | 0.99 | 1.00 |
| | | SMOTE 0.5 | 0.89 | 0.89 | 0.97 | 0.99 | 0.98 |

Syed and Hasan| | | | | | | | |
|---|---|---|---|---|---|---|---|
| | | SMOTE 1 | 0.78 | 0.78 | 0.94 | 0.99 | 0.87 |
| All | Accuracy | SMOTE 0.25 | 0.80 | 0.80 | 0.90 | 0.97 | 0.91 |
| | | SMOTE 0.5 | 0.74 | 0.74 | 0.89 | 0.96 | 0.91 |
| | | SMOTE 1 | 0.72 | 0.72 | 0.90 | 0.95 | 0.89 |

We also observe that performance of a model highly depends on the sampling strategy selected for data balancing. As the ratio of crash to non-crash events decreases, we observed a reduction in performance for all the models except XGboost which gives similar recall values across different sampling strategies. It performs better than GPBoost at SMOTE 0.5 and SMOTE 0.25. Its high accuracy and scalability promise good potential for practical implementation of the proposed model. Because of the synthetic oversampling, it is possible that most of the models may be overfitting the data which results in very high specificity and precision. With emergence of the connected vehicle data and crash record, it is likely that in the future we can empirically model the crash risk without the need for any oversampling technique.

To further validate the model results and prove the potential of connected vehicle data in crash risk prediction, we have compiled five previous studies that are relevant to our work (*Table 4*). It is important to note that each work had a different data source and sample size. Also, we only kept the best result achieved by the authors in their respective papers. Most studies utilized around 1 year of data providing more crash samples for model training and implement deep neural networks. Our shallow machine learning models have performed better than most of the previously proposed crash risk prediction models as the highest recall value of 0.91 is observed in this study compared to the highest recall value of 0.888 found from a CNN model by Cai et al. (*13*).

Table 4 : Comparison of model results from previous crash risk prediction models

| Reference | Roadway Context | Study data period | Sample size (no. of crashes) | Sampling Ratio | Model | Recall | Specificity |
|---|---|---|---|---|---|---|---|
| Yuan et al. 2019 (*47*) | Urban intersections | 1 year (2017-2018) | 665 | Did not mention | LSTM-RNN | 0.606 | |
| | | | | | Conditional Logistic Model | 0.567 | |
| Li 2020 (*58*)(*59*) | Urban arterial | 1 year (2017-2018) | 110 | 1:1 | LSTM-CNN | 0.868 | |
| | | | | | LSTM | 0.8 | |
| | | | | | CNN | 0.65 | |
| | | | | | XGBoost | 0.7 | |
| Cai et al. 2020 (*13*) | Expressways | 1 year (2017-2018) | 625 | 1:4 | Logistic Regression | 0.763 | 0.761 |
| | | | | | SVM | 0.846 | 0.800 |
| | | | | | ANN | 0.872 | 0.884 |
| | | | | | CNN | **0.888** | 0.907 |
| Basso et al. 2021(*60*) | urban highway | 1.25 (Jan 2018 - Mar 2019) | 910 | 1:1 | CNN | 0.658 | |
| | | | | | Random Forest | 0.5 | |
| | | | | | SVM sigmoid | 0.553 | |
| | | | | | Logistic Regression | 0.447 | |

Syed and Hasan

| | Freeways | 2 months (Nov 2020-Dec 2020) | 141 | 1:2 | Bi-LSTM CNN | 0.772 | |
| Zhang et al. 2022 (*30*) | | | | | LSTM + CNN | 0.723 | |

*NM means not mentioned

**CONCLUSIONS**

Predicting crash risks using real-time connected vehicle data will be a breakthrough technology in the field of traffic safety. While most crash risk prediction models rely on infrastructure-based data sources, connected vehicle data will have more coverage, potentially covering many roads that are less likely to be monitored by infrastructure sensors. In addition, instead of aggregate traffic related features such as volume and speed, vehicle data provide individual vehicle-level features such as vehicle speed and acceleration that are directly related with a crash occurrence. Finally, since these data are generated and communicated from individual vehicles, predicted crash risks can be relayed back to those individual vehicles to take precautionary measures.

In this paper, we have presented a framework of processing connected vehicle data to model real-time potential crash risks. Result from the XGBoost model shows that we can accurately predict a crash event with 91% recall rates and 0.95 f1-score. GPBoost also scored 91% recall rate for sampling ratio of 1:1. Even though GPBoost has underperformed in other sampling ratio 0.25 and 0.5, it promises to be the great fit for modelling real-time crash risks since it utilizes both boosting and gaussian processes. Overall XGBoost has also shown good performances in all sampling ratio. We also found the important features that can contribute toward crash risk prediction. Several features calculated for the previous time step and the vehicle counts (which is a proxy estimate of vehicle flow) have higher importance values. Features representing mean acceleration and mean speed also have high importance values.

This study has some limitations. Since crash events are rare, it is very difficult to train a machine learning-based crash prediction model without using any oversampling technique. The model results therefore are highly dependent on the quality of the synthetically produced data. Furthermore, we only have the data for three days, which is relatively small. Although this work demonstrates one of the futuristic applications of connected vehicle data, evacuation periods have a limited number of observations as they are typically last only few days. Thus, we had to resort to a small dataset.

Despite the shortcomings, this study suggests that connected vehicle data can be used to predict potential crash risk especially during emergency events such as hurricane evacuation. The high recall rate of the prediction model means that disaster and traffic management agencies will be able to better assess the traffic situation in advance and deploy crash counter measures more efficiently. Through this application, it is also possible to alert the mass people about impending crash risks and transmit advanced warning to drivers of any potential risk.

**ACKNOWLEDGMENTS**

This research was supported by NSF grant ID #2122135 titled as "EAGER-SAI: Exploring Pathways of Adaptive Infrastructure Management with Rapidly Intensifying Hurricanes". The authors are solely responsible for the findings presented here.

**AUTHOR CONTRIBUTIONS**

The authors confirm contribution to the paper as follows: study conception and design: Zaheen E Muktadi Syed, Samiul Hasan; data collection: Zaheen E Muktadi Syed, Samiul Hasan; analysis and interpretation of results: Zaheen E Muktadi Syed, Samiul Hasan; draft manuscript preparation and editing: Zaheen E Muktadi Syed, Samiul Hasan; funding acquisition and supervision: Samiul Hasan. All authors reviewed the results and approved the final version of the manuscript.

Syed and Hasan16. Af Wåhlberg, A. E. The Stability of Driver Acceleration Behavior, and a Replication of Its Relation to Bus Accidents. *Accident Analysis and Prevention*, Vol. 36, No. 1, 2004, pp. 83–92. https://doi.org/10.1016/S0001-4575(02)00130-6.
17. Jun, J., Ogle, J., & Guensler, R. (2007). Relationships between Crash Involvement and Temporal-Spatial Driving Behavior Activity Patterns: Use of Data for Vehicles with Global Positioning Systems. *Transportation Research Record,* 2019(1), 246–255. https://doi.org/10.3141/2019-29.
18. Xie, K., X. Wang, H. Huang, and X. Chen. Corridor-Level Signalized Intersection Safety Analysis in Shanghai, China Using Bayesian Hierarchical Models. *Accident Analysis & Prevention*, Vol. 50, 2013, pp. 25–33. https://doi.org/10.1016/J.AAP.2012.10.003.
19. Stipancic, Joshua & Miranda-Moreno, Luis & Saunier, Nicolas & Labbe, Aurelie. (2018). Surrogate Safety and Network Screening: Modelling Crash Frequency Using GPS Data and Latent Gaussian Models. Accident; analysis and prevention. 120. 174-187. 10.1016/j.aap.2018.07.013.
20. Kamrani, M., R. Arvin, and A. J. Khattak. The Role of Aggressive Driving and Speeding in Road Safety: Insights from SHRP2 Naturalistic Driving Study Data. 2019. Transportation Research Board 98th Annual Meeting.
21. Desai, J., H. Li, J. K. Mathew, Y.-T. Cheng, A. Habib, and D. M. Bullock. Correlating Hard-Braking Activity with Crash Occurrences on Interstate Construction Projects in Indiana. *Journal of Big Data Analytics in Transportation 2020 3:1*, Vol. 3, No. 1, 2020, pp. 27–41. https://doi.org/10.1007/S42421-020-00024-X.
22. Hunter, M., E. Saldivar-Carranza, J. Desai, J. K. Mathew, H. Li, and D. M. Bullock. A Proactive Approach to Evaluating Intersection Safety Using Hard-Braking Data. *Journal of Big Data Analytics in Transportation*, Vol. 3, 2021, pp. 81–94. https://doi.org/10.1007/s42421-021-00039-y.
23. Yu, R., and M. Abdel-Aty. Utilizing Support Vector Machine in Real-Time Crash Risk Evaluation. *Accident Analysis & Prevention*, Vol. 51, 2013, pp. 252–259. https://doi.org/10.1016/J.AAP.2012.11.027.
24. Wang, L., Q. Shi, and M. Abdel-Aty. Predicting Crashes on Expressway Ramps with Real-Time Traffic and Weather Data: *Transportation Research Record, 2514(1), 32–38.*, Vol. 2514, 2015, pp. 32–38. https://doi.org/10.3141/2514-04.
25. Gan, J., L. Li, D. Zhang, Z. Yi, and Q. Xiang. An Alternative Method for Traffic Accident Severity Prediction: Using Deep Forests Algorithm. *Journal of Advanced Transportation*, Vol. 2020, 2020. https://doi.org/10.1155/2020/1257627.
26. Shi, X., Y. D. Wong, M. Z. F. Li, C. Palanisamy, and C. Chai. A Feature Learning Approach Based on XGBoost for Driving Assessment and Risk Prediction. *Accident Analysis & Prevention*, Vol. 129, 2019, pp. 170–179. https://doi.org/10.1016/J.AAP.2019.05.005.
27. Li-Li Wang, Henry Y.T. Ngan, Nelson H.C. Yung, Automatic incident classification for large-scale traffic data by adaptive boosting SVM. Information Sciences, Volume 467, 2018, Pages 59-73, ISSN 0020-0255, https://doi.org/10.1016/j.ins.2018.07.044.
28. Huang, T., S. Wang, and A. Sharma. Highway Crash Detection and Risk Estimation Using Deep Learning. *Accident Analysis & Prevention*, Vol. 135, 2020, p. 105392. https://doi.org/10.1016/J.AAP.2019.105392.
29. Li, P., M. Abdel-Aty, and J. Yuan. Real-Time Crash Risk Prediction on Arterials Based on LSTM-CNN. *Accident Analysis and Prevention*, Vol. 135, 2020. https://doi.org/10.1016/J.AAP.2019.105371.
30. Zhang, S., and M. Abdel-Aty. Real-Time Crash Potential Prediction on Freeways Using Connected Vehicle Data. *Analytic Methods in Accident Research*, Vol. 36, 2022, p. 100239. https://doi.org/10.1016/J.AMAR.2022.100239.
31. Xu, C., W. Wang, P. Liu, and F. Zhang. Development of a Real-Time Crash Risk Prediction Model Incorporating the Various Crash Mechanisms Across Different Traffic States. *Traffic Injury Prevention*, Vol. 16, No. 1, 2015, pp. 28–35. https://doi.org/10.1080/15389588.2014.909036.
23